\documentclass[conference]{IEEEtran}
\IEEEoverridecommandlockouts

\usepackage{cite}
\usepackage{amsmath,amssymb,amsfonts}
\usepackage{graphicx}
\usepackage{textcomp}
\usepackage{xcolor}
\usepackage{booktabs}
\usepackage{multirow}
\usepackage{hyperref}
\usepackage{url}

\usepackage{subfigure}
\usepackage{amsthm}
\def\BibTeX{{\rm B\kern-.05em{\sc i\kern-.025em b}\kern-.08em
    T\kern-.1667em\lower.7ex\hbox{E}\kern-.125emX}}
\begin{document}

\title{Beyond Phenotypic Similarity: Relation Refinement and Selective Population Graph Learning for Multimodal Brain Disorder Diagnosis}

\author{\IEEEauthorblockN{Yinyu Lan}
\IEEEauthorblockA{School of Medicine and Health, Harbin Institute of Technology\\
Harbin 150001, China\\
lanyinyu19@mails.ucas.ac.cn}
}



\maketitle

\begin{abstract}
Population graph learning supports brain disorder diagnosis by sharing information across subjects. Its effectiveness, however, depends on whether the relationships guiding this exchange are relevant to diagnosis. Similarity in demographic or acquisition attributes does not necessarily indicate diagnostic agreement, and heterogeneous subjects may require different representation updates within the same population graph. We propose PopDRP, a framework that combines diagnosis-guided relation refinement with hierarchical selective state-space learning. Using only training labels, Contrastive Affinity Refinement calibrates attribute-based affinities according to diagnostic agreement. Neural Graph Rectification then refines these relationships using fused functional-connectivity and non-imaging representations, guided by graph reconstruction and dual-view alignment objectives. On the resulting graph, a Selective Pyramid State-Space backbone integrates input-dependent state updates, graph diffusion, and hierarchical pooling and unpooling to learn population representations at multiple resolutions. Together, these components connect the refinement of inter-subject relationships with selective representation learning over the population graph. Experiments on ABIDE and ADHD-200 under stratified ten-fold cross-validation show improved accuracy and AUC over the compared baselines. Ablation studies further support the contributions of relation refinement and the selective hierarchical backbone to classification performance.
\end{abstract}


\section{Introduction}
\label{sec:intro}

Population graph learning provides a way to incorporate inter-subject information into brain disorder diagnosis \cite{parisot2017spectral,parisot2018disease}. In this formulation, each node represents a subject, node features describe individual measurements such as resting-state functional connectivity, and edges determine how information is shared across the cohort. Non-imaging attributes, including age, sex, and acquisition site, provide additional cues for constructing these relationships \cite{parisot2018disease,wang2023evgcn}. The value of a population graph therefore depends not only on the quality of individual features, but also on whether the relationships connecting subjects support the classification task.

A central modeling difficulty is that \emph{attribute similarity and diagnostic relevance are not interchangeable}. Two subjects may share demographic characteristics or an acquisition site while having different diagnoses. Attribute-based affinities consequently provide useful relational priors, but their implications for information sharing must be considered in relation to the task. Existing research has already moved beyond fixed graph construction: EV-GCN learns adaptive inter-subject associations \cite{wang2023evgcn}, while MM-GTUNets incorporates label-informed affinity modeling and hierarchical graph encoding \cite{yao2024mmgtunets}. Building on these advances, we investigate how diagnosis-guided affinities can be further refined using multimodal representations and incorporated into a selective hierarchical population encoder.

\begin{figure}
    \centering
    \includegraphics[width=1\linewidth]{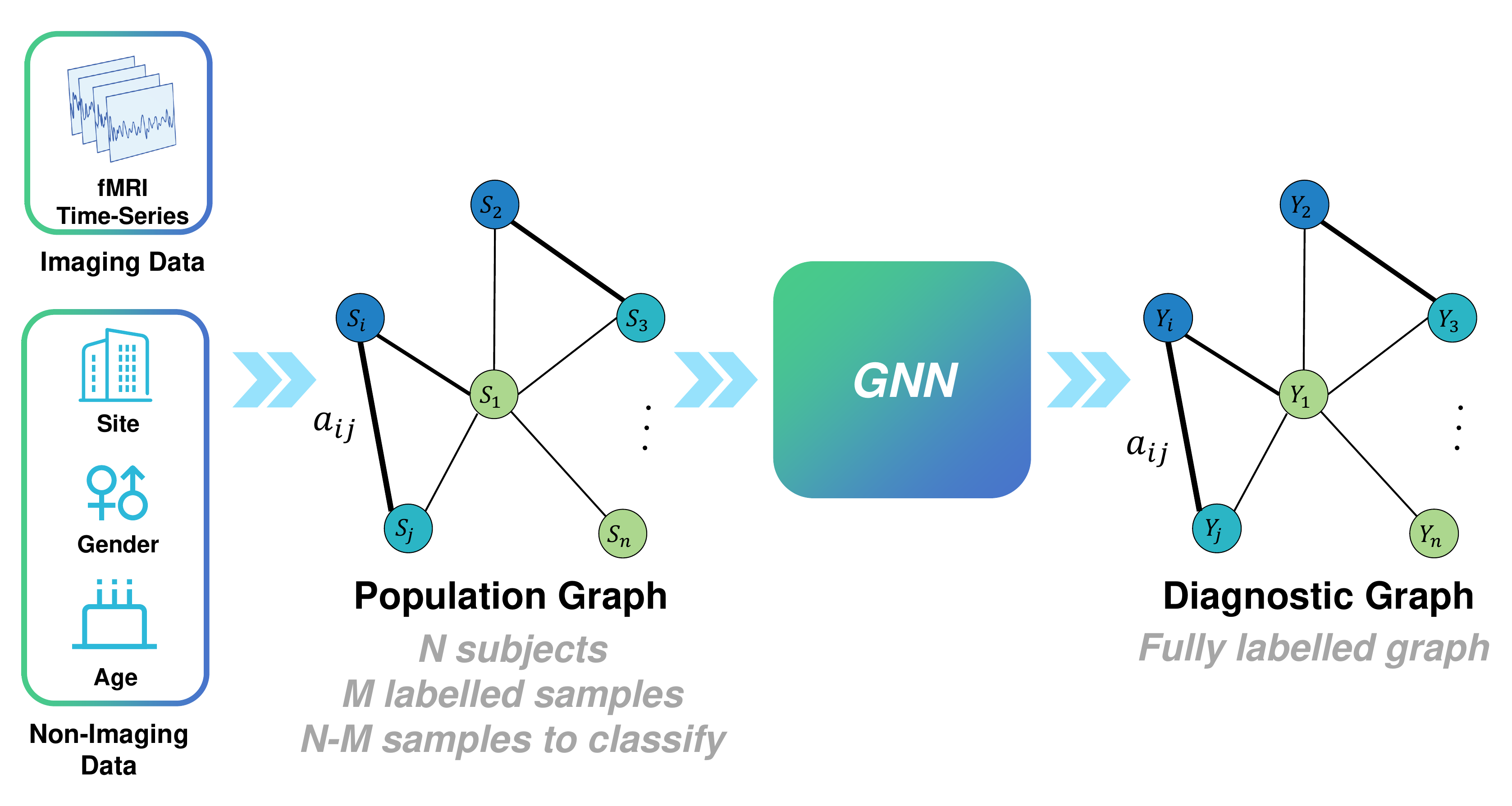}
    \caption{Multimodal brain disease diagnosis based on population graph.}
    \label{fig:spss}
\end{figure}

Relation refinement and population representation learning are closely connected. Edge weights determine which subjects influence one another, while the encoder determines how the resulting information is transformed. A refined graph provides a basis for information exchange, but heterogeneous subject features still motivate input-dependent representation updates. Selective state-space models offer a mechanism for conditioning state transformations on the input, and their adaptation to graph learning has motivated investigation into graph-aware information processing \cite{gu2023mamba,wang2024graphmamba,behrouz2024mambaw}. Hierarchical graph processing introduces a further consideration: representations should integrate information at different population resolutions while retaining features needed for individual predictions \cite{yao2024mmgtunets, guo2022}. These considerations motivate a framework that combines feature-conditioned relation refinement with selective state updates throughout a graph hierarchy.

We present \textbf{PopDRP} to implement this design through three connected stages. First, Contrastive Affinity Refinement (CAR) uses diagnostic agreement among training subjects to calibrate non-imaging affinities, producing a diagnosis-guided population prior. Second, Neural Graph Rectification (NGR) combines this prior with fused imaging and non-imaging features to learn refined edge weights. Graph reconstruction and dual-view alignment constrain this process. Third, the Selective Pyramid State-Space (SPSS) backbone processes the refined graph through input-dependent state updates and topology-aware diffusion within a pooling and unpooling hierarchy. Skip connections retain subject-level information as graph resolution changes. The resulting architecture links the refinement of population relationships to selective representation learning across the hierarchy.

The contributions of this work are threefold:
\begin{itemize}
    \item We develop a population graph refinement pipeline that combines training-label-informed affinities with feature-conditioned edge learning. This allows multimodal subject representations to refine the relationships used for downstream classification.

    \item We introduce a Selective Pyramid State-Space backbone that integrates input-dependent state updates and graph diffusion within a hierarchical encoder--decoder. It learns population representations at multiple resolutions while retaining information for subject-level prediction.

    \item We evaluate PopDRP on ABIDE and ADHD-200 using stratified ten-fold cross-validation. Comparisons with established baselines and component ablations assess the effectiveness of the framework and the contributions of its principal components.
\end{itemize}

\begin{figure*}[!ht]
    \centering
    \includegraphics[width=1\linewidth]{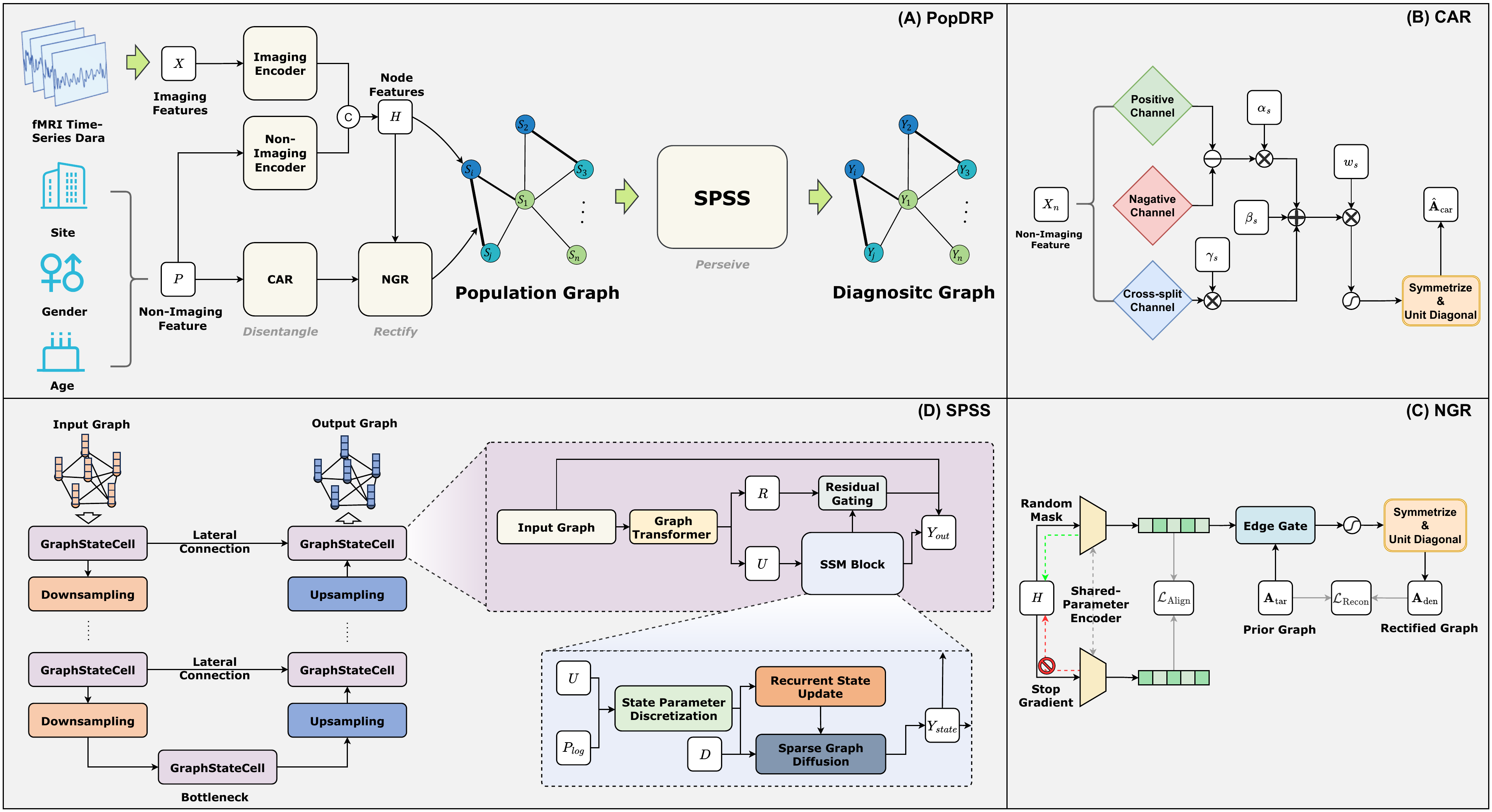}
    \caption{(A) The framework of PopDRP. (B) Contrastive Affinity Refinement (CAR) module. (C) Neural Graph Rectification (NGR) module. (D) Architecture of the Selective Pyramid State-Space (SPSS).}
    \label{fig:spss}
\end{figure*}
\section{Related Work}\label{sec:related}
\subsection{Population Graph-Based Brain Disease Diagnosis}
Parisot et al. pioneered population graph learning for brain disease diagnosis via handcrafted non-imaging similarity and GCN-based classification \cite{parisot2017spectral, parisot2018disease}. Follow-up works optimized this framework via multi-scale graph convolution \cite{kazi2019inception}, ROI-aware graph construction \cite{li2022braingnn}, multimodal fusion, graph transformer and adversarial contrastive learning \cite{wang2023evgcn,yao2024mmgl,yao2024mmgtunets,zheng2023agcl}. Even so, most methods build population graphs with fixed handcrafted similarity metrics on raw non-imaging data, suffering severely from missing data, site effects and label noise. 

\subsection{State-Space Models and Graph Mamba}
SSMs outperform attention models for long-range sequence modeling with linear computational complexity. Classic S4 and Mamba capture long-range dependencies via continuous latent states and input-selective gating \cite{gu2022efficiently,gu2023mamba}. Extended graph Mamba methods serialize graph nodes into sequences to fit Mamba blocks for graph learning \cite{wang2024graphmamba,behrouz2024mambaw}, yet they only support single-scale modeling without hierarchical graph pooling. To tackle this issue, our SPSS backbone embeds selective SSM units into a U-Net pyramid structure and adopts graph diffusion for topology-aware state propagation, supporting joint multi-scale and long-range representation learning for population graphs.

\subsection{Graph Denoising and Structure Learning}
Graph structure learning jointly optimizes graph topology and node embeddings \cite{zhu2021deep}. Existing graph rectification methods refine graph edges relying on node features or inherent topology \cite{song2022tam,song2023rge}, which are tailored for homogeneous graphs and fail to introduce diagnostic label supervision. Besides, population graphs face unique noise from defective non-imaging attributes and site confounding factors. Our NGR module addresses this gap by modeling edge feature interactions and leveraging clinical label priors from CAR-optimized graphs under a minimum-sufficient-representation objective.

\section{Method}\label{sec:method}

\subsection{Problem Formulation}\label{sec:formulation}

Let $\mathcal{D} = {(\mathbf{X}_i, \mathbf{p}i, y_i)}{i=1}^{N}$ be a cohort of $N$ subjects, where $\mathbf{X}_i \in \mathbb{R}^{D}$ is the vectorized upper-triangular functional connectivity feature, $\mathbf{p}_i \in \mathbb{R}^{S}$ denotes non-imaging attributes (e.g., age, gender, and site), and $y_i \in {0,1}$ is the diagnostic label. We first construct a population graph $\mathcal{G} = (\mathcal{V}, \mathcal{E}, \mathbf{A})$ from multimodal imaging and non-imaging information, where each subject is represented as a node and edges encode inter-subject relationships. Brain disease diagnosis is then formulated as node classification over the learned population graph, i.e.,$(\mathbf{X}, \mathbf{P}) \rightarrow \mathcal{G} \rightarrow \hat{\mathbf{y}}$.

\subsection{Overview of PopDRP}

PopDRP consists of four stages: (1) multimodal preprocessing (RFE for imaging, VAE for non-imaging), (2) Contrastive Affinity Refinement (CAR) to build a clinically informed graph, (3) Neural Graph Rectification (NGR) to denoise the graph via minimum-sufficient representation, and (4) Selective Pyramid State-Space (SPSS) backbone for hierarchical representation, followed by MLP classification.

\subsection{Multimodal Feature Preprocessing}

\subsubsection{Imaging Feature Encoding}

We vectorize the upper-triangular part of each connectivity matrix and apply recursive feature elimination (RFE) with a ridge classifier to select the top $D$ discriminative features using only training data:
\begin{equation}
\mathbf{X} = \text{RFE}\left(\mathbf{X}_{\text{raw}}, \mathbf{y}_{\text{train}}, D\right), \quad \mathbf{X} \in \mathbb{R}^{N \times D}.
\end{equation}

\subsubsection{Phenotypic Feature Encoding}

A variational autoencoder (VAE) is trained unsupervised on $\mathbf{P} \in \mathbb{R}^{N \times S}$. The encoder produces $\boldsymbol{\mu}_i, \log \boldsymbol{\sigma}_i^2 = \text{MLP}_{\text{enc}}(\mathbf{p}_i)$, and we sample $\mathbf{z}_i = \boldsymbol{\mu}_i + \boldsymbol{\sigma}_i \odot \boldsymbol{\epsilon}$ via reparameterization. The decoder $\hat{\mathbf{p}}_i = \text{MLP}_{\text{dec}}(\mathbf{z}_i)$ reconstructs the input. The VAE loss is:
\begin{equation}
\mathcal{L}_{\text{VAE}} = \frac{1}{N}\sum_{i=1}^{N}\|\hat{\mathbf{p}}_i - \mathbf{p}_i\|_2^2 + \beta \cdot D_{\text{KL}}\left(\mathcal{N}(\boldsymbol{\mu}_i, \boldsymbol{\sigma}_i^2) \| \mathcal{N}(\mathbf{0}, \mathbf{I})\right).
\end{equation}
After pretraining, latent embeddings $\mathbf{Z} = [\mathbf{z}_1, \dots, \mathbf{z}_N]^\top \in \mathbb{R}^{N \times D}$ serve as denoised non-imaging features; the encoder is frozen thereafter.

\subsection{Contrastive Affinity Refinement (CAR)}

For each non-imaging attribute $s$, we decompose pairwise similarity into three matrices using only training labels. For continuous attributes (threshold $t$):
\begin{align}
\mathbf{P}^{(s)}_{ij}&=\mathbf{1}[i,j\in\mathcal{V}_{\text{train}}, |x_i^{(s)}-x_j^{(s)}|<t, y_i=y_j],\\
\mathbf{N}^{(s)}_{ij}&=\mathbf{1}[i,j\in\mathcal{V}_{\text{train}}, |x_i^{(s)}-x_j^{(s)}|<t, y_i\neq y_j],\\
\mathbf{C}^{(s)}_{ij}&=\mathbf{1}[i,j\in\mathcal{V}_{\text{val}}\cup\mathcal{V}_{\text{test}}, |x_i^{(s)}-x_j^{(s)}|<t].
\end{align}
For categorical attributes, replace $|x_i^{(s)}-x_j^{(s)}|<t$ with $x_i^{(s)}=x_j^{(s)}$. CAR computes a refined affinity per attribute:
\begin{equation}
\tilde{\mathbf{A}}^{(s)} = \alpha_s (\mathbf{P}^{(s)} - \mathbf{N}^{(s)}) + \beta_s + \gamma_s \mathbf{C}^{(s)},
\end{equation}
with $\alpha_s,\gamma_s = \text{Softplus}(\cdot) > 0$. Learnable importance weights $w_s = \text{softmax}(\hat{w}_s)$ aggregate:
\begin{equation}
\mathbf{A}_{\text{car}} = \sigma\left(\frac{1}{\tau}\sum_{s=1}^{S} w_s \tilde{\mathbf{A}}^{(s)}\right), \quad \tau = |\hat\tau|>0.
\end{equation}
The final symmetric adjacency with self-loops is:

\begin{equation}
\hat{\mathbf{A}}_\text{car} = \frac12(\mathbf{A}_{\text{car}}+\mathbf{A}_{\text{car}}^\top), \hat{\mathbf{A}}_\text{car}[i,i]=1.
\end{equation}

\subsection{Graph Construction and Neural Graph Rectification (NGR)}

We construct fused node features $\mathbf{H} = [\mathbf{X}, \mathbf{Z}] \in \mathbb{R}^{N \times 2D}$. A Gaussian RBF similarity graph is:
\begin{equation}
\mathbf{A}_{\text{sim}} = \exp\left(-\frac{\|\mathbf{h}_i - \mathbf{h}_j\|_2^2}{2\sigma^2}\right).
\end{equation}
The target graph combines similarity and clinical prior: $\mathbf{A}_{\text{tar}} = \mathbf{A}_{\text{sim}} \odot \hat{\mathbf{A}}_\text{car}$. NGR learns a minimum sufficient representation $\mathbf{U}=f(\mathbf{H})$ by optimizing:

\begin{equation}
\mathcal{L}_{\text{graph}} = \underbrace{\text{MSE}(\mathbf{A}_{\text{den}}, \mathbf{A}_{\text{tar}})}_{\mathcal{L}_{\text{recon}} } + \underbrace{\left(1 - \frac{1}{N}\sum_{i=1}^{N} \frac{\mathbf{U}_i \cdot \mathbf{U}_{\text{ref},i}}{\|\mathbf{U}_i\| \|\mathbf{U}_{\text{ref},i}\|}\right)}_{\mathcal{L}_{\text{align}}},
\label{eq:graphloss}
\end{equation}

where $\mathbf{U}_{\text{ref}}$ is a second encoding with detached gradients. The rectified graph $\mathbf{A}_{\text{den}}$ is predicted from $\mathbf{U}$ via:
\begin{equation}
\mathbf{A}_{\text{den}}^{ij} = \sigma\!\left(\text{MLP}_{\text{edge}}\big([\mathbf{U}_i; \mathbf{U}_j; |\mathbf{U}_i-\mathbf{U}_j|; \mathbf{U}_i\odot\mathbf{U}_j; \mathbf{A}_{\text{tar},ij}]\big)\right),
\end{equation}
then symmetrized, masked by $\mathbf{A}_{\text{tar}}$, and filled with self-loops.

\subsection{Selective Pyramid State-Space (SPSS)}
The rectified graph $\mathbf{A}_{\text{den}}$ serves as computation basis for SPSS, the core module of PopDRP. SPSS is designed following three neuroscience-derived principles: local message passing captures within-subsystem brain region interactions; selective state-space recurrence models cross-population long-range sequential dependencies; hierarchical pyramid pooling/unpooling constructs multi-scale representations matching hierarchical human brain functional organization.

\subsubsection{SPSS Architecture}
SPSS adopts a U-Net structure with $L$ encoder layers, one bottleneck layer, and $L$ decoder layers. Denote input node feature $\mathbf{H}^{(0)} = \mathbf{H}$ and original graph $\mathcal{G}^{(0)} = (\mathcal{V}^{(0)}, \mathcal{E}^{(0)})$ built on $\mathbf{A}_{\text{den}}$. For each encoder layer $\ell = 0, \dots, L-1$, node features are updated via GraphStateCell before pooling:
\begin{equation}
\mathbf{H}^{(\ell)} \leftarrow \text{GraphStateCell}_{\text{down}}^{(\ell)}\left(\mathbf{H}^{(\ell)}, \mathcal{G}^{(\ell)}\right), 
\end{equation}
\begin{equation}
\mathbf{H}^{(\ell+1)}, \mathcal{G}^{(\ell+1)}, \mathbf{P}^{(\ell)} \leftarrow \text{TopKPool}\left(\mathbf{H}^{(\ell)}, \mathcal{G}^{(\ell)}\right),
\end{equation}
where $\mathbf{P}^{(\ell)}$ stores indices of preserved nodes. TopKPool retains the top-$r$ ratio nodes by projection score to allocate computation to informative nodes. After $L$ pooling steps, the coarsened graph contains global population structural information, processed by bottleneck GraphStateCell:
\begin{equation}
\mathbf{H}^{(L)} \leftarrow \text{GraphStateCell}_{\text{bottleneck}}\left(\mathbf{H}^{(L)}, \mathcal{G}^{(L)}\right).
\end{equation}

Decoders restore graph resolution via unpooling and lateral skip GraphStateCell fusion:
\begin{align}
\mathbf{H}_{\text{up}}^{(\ell)} &\leftarrow \text{Unpool}\left(\mathbf{H}^{(\ell+1)}, \mathbf{P}^{(\ell)}\right), \\
\mathbf{H}^{(\ell)} &\leftarrow \text{GraphStateCell}_{\text{lateral}}^{(\ell)}\left(\mathbf{H}^{(\ell)}, \mathcal{G}^{(\ell)}\right), \\
\mathbf{H}^{(\ell)} &\leftarrow \text{GraphStateCell}_{\text{up}}^{(\ell)}\left(\mathbf{H}_{\text{up}}^{(\ell)} + \mathbf{H}^{(\ell)}, \mathcal{G}^{(\ell)}\right).
\end{align}
Unpool maps coarse features back to original node positions with zero padding for pruned nodes; lateral skip connections recover fine-grained local structures lost during pooling.

\subsubsection{Selective State-Space with Graph Diffusion (GraphStateCell)}
GraphStateCell is SPSS’s fundamental computation unit, integrating graph attention convolution, input-selective SSM recurrence, and sparse graph diffusion to align hidden states with graph topology. Layer-normalized input first passes a TransformerConv + ReLU block:
\begin{equation}
\mathbf{H}_{\text{conv}} = \text{ReLU}\left(\text{TransformerConv}\left(\text{LayerNorm}(\mathbf{H}), \mathbf{A}_{\text{den}}\right)\right).
\end{equation}

\paragraph{State-space projection} Convolution outputs are linearly split into SSM input branch $\mathbf{U}$ and residual branch $\mathbf{R}$:
\begin{equation}
[\mathbf{U}, \mathbf{R}] = \mathbf{H}_{\text{conv}} \mathbf{W}_{\text{in}}, \quad \mathbf{U}, \mathbf{R} \in \mathbb{R}^{N \times d_{\text{inner}}}.
\end{equation}

\paragraph{State Parameter Discretization} Unlike conventional fixed-parameter SSMs, SPSS generates input-adaptive transition parameters $\Delta, \mathbf{B}, \mathbf{C}$:
\begin{equation}
[\Delta', \mathbf{B}, \mathbf{C}] = \mathbf{U} \mathbf{W}_x, \quad \Delta = \text{Softplus}(\Delta' \mathbf{W}_{\Delta}).
\end{equation}
Diagonal state matrix $\mathbf{A} = -\exp(\mathbf{A}_{\log})$ uses learnable weights to stabilize long-range state propagation.

\paragraph{Recurrent State Update} Hidden states are iteratively updated for $K$ steps with zero initialization $\mathbf{S}_0 = \mathbf{0}$:
\begin{equation}
\mathbf{S}_k = \exp(\Delta \otimes \mathbf{A}) \odot \mathbf{S}_{k-1} + \mathbf{U} \otimes \mathbf{B},
\end{equation}
where $\otimes$ denotes outer product and $\odot$ element-wise multiplication. Iterative refinement accumulates multi-step contextual information per node.

\paragraph{Sparse Graph Diffusion} Post-recurrence hidden states diffuse over graph edges for neighborhood aggregation:
\begin{equation}
\tilde{\mathbf{S}}_K = \tilde{\mathbf{A}}_{\text{sym}} \cdot \text{flatten}(\mathbf{S}_K),
\end{equation}
with symmetric normalized adjacency matrix $\tilde{\mathbf{A}}_{\text{sym}} = \mathbf{D}^{-1/2}(\mathbf{A}_{\text{den}} + \mathbf{I})\mathbf{D}^{-1/2}$. Sparse matrix multiplication avoids $\mathcal{O}(N^2)$ dense memory cost. Preliminary SSM output fuses diffused states, projection $\mathbf{C}$ and channel bias $\mathbf{D}$:
\begin{equation}
\mathbf{Y}_{\text{ssm}} = (\tilde{\mathbf{S}}_K \odot \mathbf{C}) + \mathbf{U} \cdot \mathbf{D},
\end{equation}
$\mathbf{D} \in \mathbb{R}^{d_{\text{inner}}}$ is learnable channel bias.

\paragraph{Residual gating} A temperature gate balances long-range SSM and local residual signals:
\begin{equation}
\mathbf{G} = \sigma\left(\frac{[\mathbf{Y}_{\text{ssm}}, \mathbf{R}] \mathbf{W}_{\text{gate}}}{\tau_g}\right), 
\end{equation}
\begin{equation}
\mathbf{Y}_{\text{out}} = \left(\mathbf{Y}_{\text{ssm}} + \mathbf{G} \odot \text{ReLU}(\mathbf{R})\right) \mathbf{W}_{\text{out}}.
\end{equation}
Standard residual connection with dropout yields final cell output:
\begin{equation}
\mathbf{Y}_{\text{cell}} = \mathbf{H} + \text{Dropout}(\mathbf{Y}_{\text{out}}).
\end{equation}

Conventional SSMs use shared fixed $(\Delta, \mathbf{A}, \mathbf{B}, \mathbf{C})$ for all samples, leading to uniform feature compression that fails to model heterogeneous brain connectivity across neurodevelopmental disorder subjects. SPSS dynamically computes parameters per input, enabling subject-adaptive feature selection for diverse multimodal brain graphs.

\subsection{Classification and Training Objective}
SPSS output embeddings $\mathbf{H}_{\text{final}} \in \mathbb{R}^{N \times d_{\text{out}}}$ feed a two-layer BN-ReLU MLP classifier:
\begin{equation}
\hat{\mathbf{y}} = \operatorname{Softmax}\left(\mathbf{W}_2 \cdot \operatorname{BN}\left(\operatorname{ReLU}\left(\mathbf{W}_1 \mathbf{H}_{\text{final}} + \mathbf{b}_1\right)\right) + \mathbf{b}_2\right).
\end{equation}

Total training loss combines supervised cross-entropy and graph smoothness regularization:
\begin{equation}
\mathcal{L}_{\text{total}} = -\frac{1}{|\mathcal{I}_{\text{train}}|}\sum_{i \in \mathcal{I}_{\text{train}}} \sum_{c=0}^{1} \mathbb{I}[y_i = c] \log \hat{y}_{i,c} + \lambda_{\text{graph}} \cdot \mathcal{L}_{\text{graph}},
\end{equation}
where $\mathcal{L}_{\text{graph}}$ follows Eq.~(\ref{eq:graphloss}), and $\lambda_{\text{graph}}$ weights regularization strength. All model parameters are end-to-end optimized via Adam.

\section{Experiments}\label{sec:experiments}

\subsection{Datasets and Preprocessing}

We evaluate PopDRP on two widely used public neurodevelopmental disorder datasets.

\textbf{ABIDE Dataset:} The Autism Brain Imaging Data Exchange (ABIDE) dataset \cite{di2014autism} aggregates rs-fMRI and non-imaging data from 17 international imaging sites. Following standard preprocessing with the Configurable Pipeline for the Analysis of Connectomes (CPAC), we obtain AAL atlas-based region-averaged time series for 116 regions of interest (ROIs) from 871 subjects (468 healthy controls, 403 individuals with ASD). Functional connectivity matrices are computed as Pearson correlation coefficients between ROI time series, followed by Fisher-Z transformation. Age, gender, and scanning site are used as non-imaging attributes.

\textbf{ADHD-200 Dataset: }The ADHD-200 consortium dataset \cite{bellec2017adhd200} provides rs-fMRI from 582 subjects (364 healthy controls, 218 individuals with ADHD) across multiple sites. Data are preprocessed using the Athena pipeline with nuisance signal removal. Functional connectivity matrices are computed identically to ABIDE using the AAL atlas. Age, gender, and site serve as non-imaging attributes.



For both datasets, we apply Fisher-Z normalization to connectivity values, extract the upper triangular portion of each $116 \times 116$ connectivity matrix as a 6670-dimensional feature vector, and apply RFE to select the $D = 500$ most discriminative features. Stratified 10-fold cross-validation is employed, with an additional 10\% of training data held out as a validation set for each fold.

\subsection{Evaluation Metrics and Baselines}

We evaluate performance using five metrics: classification accuracy (ACC), sensitivity (SEN), specificity (SPE), area under the ROC curve (AUC), and F1-score. We compare PopDRP against three categories of methods:

We compare PopDRP against three categories of baselines. Traditional machine learning methods: SVM with RBF kernel and MLP applied to the RFE-selected connectivity features. Brain-graph-based methods: Brain-GNN \cite{li2022braingnn}, DGCN \cite{cui2022dgcn}, AL-NEGAT \cite{chen2022alnegat}, and A-GCL \cite{zheng2023agcl}, which construct individual-level brain graphs and apply GNNs for classification. Population-graph-based methods: Pop-GCN \cite{parisot2018disease}, GATE \cite{salehi2020gate}, EV-GCN \cite{wang2023evgcn}, MMGL \cite{yao2024mmgl}, and MM-GTUNets \cite{yao2024mmgtunets}, which model inter-subject relationships as population graphs. To ensure consistency, the same data splits and evaluation protocol are adopted.

\begin{table}[!t]
  \centering
  \caption{Hyperparameter settings for the ABIDE and ADHD-200 datasets.}
  \label{tab:Hyperparameter}
  \begin{tabular}{l c c}
    \toprule
    Hyperparameter & ADHD-200 & ABIDE \\
    \midrule
    SPSS Depth & 4 & 1 \\
    State dimension & 64 & 64 \\
    Mask rate & 0.2 & 0.8 \\
    Hidden dimension & 64 & 512 \\
    Learning rate & $10^{-4}$ & $10^{-4}$ \\
    Edge dropout & 0.3 & 0.3 \\
    Weight decay  & $5 \times 10^{-4}$ & $5 \times 10^{-4}$ \\
    Epoch  &  500 & 500 \\
    \bottomrule
  \end{tabular}
\end{table}

\subsection{Implementation Details}

\begin{table*}[!ht]
  \centering
  \caption{Performance comparison with three categories of baselines on ABIDE and ADHD-200 datasets (\%). The best results are marked in bold and the second-best are underlined.}
  \label{tab:performance}
  \resizebox{\linewidth}{!}{
  \begin{tabular}{cc cccc cccc}
    \toprule
    \multirow{2}{*}{Category} & \multirow{2}{*}{Method} & \multicolumn{4}{c}{ABIDE} & \multicolumn{4}{c}{ADHD-200} \\
    \cmidrule(lr){3-6} \cmidrule(lr){7-10}
    & & ACC & SEN & SPE & AUC & ACC & SEN & SPE & AUC \\
    \midrule
    \multirow{2}{*}{Traditional ML}
    & SVM & 63.21 (0.25) & 57.05 (1.05) & 68.50 (1.55) & 67.43 (1.16) & 64.67 (0.92) & 33.95 (2.18) & 83.05 (0.65) & 66.30 (0.26) \\
    & MLP & 63.71 (0.36)  & 56.28 (0.40) & 70.06 (0.35) & 68.48 (0.25) & 65.30 (0.35) & 40.30 (3.55) & 80.20 (0.45) & 67.33 (0.42) \\
     \midrule
    \multirow{4}{*}{Brain-graph}
    & Brain-GNN & 69.76 (3.80) & 67.47 (3.10) & 73.28 (3.26) & 72.50 (3.10) & 65.26 (3.60) & 68.59 (3.30) & 63.05 (4.00) & 66.02 (5.50) \\
    & DGCN & 71.83 (2.90) & 70.90 (2.59) & 71.80 (3.10) & 72.45 (2.98) & 68.81 (3.69) & 69.08 (4.07) & 69.53 (4.71) & 70.05 (4.57) \\
    & AL-NEGAT & 73.17 (3.32) & 78.18 (2.66) & 73.28 (3.14) & 75.02 (2.56) & 69.15 (3.82) & 69.28 (4.32) & 72.26 (3.92) & 68.25 (3.69) \\
    & A-GCL & 79.04 (2.40) & 81.42 (3.03) & {80.95 (3.19)} & {82.86 (2.91)} & 80.11 (4.30) & \underline{82.04 (4.58)} & 80.08 (4.10) & 78.78 (4.39) \\
   \midrule
    \multirow{5}{*}{Population-graph}
    & Pop-GCN & 68.43 (0.86) & 78.32 (0.89) & 57.51 (5.99) & 73.90 (3.08) & 75.45 (0.32) & 50.99 (1.96) & {90.11 (0.22)} & 81.72 (1.40) \\
    & GATE & 74.65 (2.50) & 75.59 (2.43) & 76.87 (2.27) & 76.87 (2.27) & 72.20 (0.23) & 77.27 (5.20) & 72.40 (3.78) & 74.61 (3.30) \\
    & EV-GCN & {80.95 (0.27)} & {83.74 (0.61)} & 77.69 (0.18) & 82.37 (0.29) & {80.95 (0.96)} & 60.33 (7.64) & \textbf{93.11 (0.13)} & 87.15 (1.15) \\
    & MMGL & 80.34 (0.18) & 80.25 (0.21) & 77.16 (3.15) & 79.83 (1.06) & 77.59 (3.64) & 76.21 (2.98) & 74.57 (1.25) & 78.47 (2.29) \\
    & MM-GTUNets & \underline{82.92 (0.54)} & \underline{84.22 (0.81)} & \underline{81.43 (0.64)} & \underline{88.21 (0.61)} & \underline{82.68 (0.60)} & {77.58 (2.11)} & \underline{85.76 (0.46)} & \underline{90.71 (0.72)} \\
    \midrule
   Ours & PopDRP & \textbf{84.18 (0.66)} & \textbf{89.14 (0.44)} & \textbf{81.59 (1.30)} & \textbf{90.55 (0.61)} & \textbf{85.79 (1.11)} & \textbf{89.96 (1.56)} & 84.46 (1.10) & \textbf{92.37 (1.22)} \\
    \bottomrule
  \end{tabular}
  }
\end{table*}

The VAE used for non-imaging feature encoding includes a two-layer MLP encoder and a two-layer MLP decoder. We pre-train this VAE for a maximum of 3000 epochs and adopt early stopping with a patience value of 100. The CAR module applies learnable affine transformations $\alpha_s, \beta_s, \gamma_s$ to each attribute across three contrastive decomposition channels for positive, negative and cross-split signals, while the learnable edge temperature $\tau$ is initialized to 1.0. Neural Graph Rectification (NGR) uses a shared two-layer node encoder to produce dual-view representations, along with a two-layer edge decoder. GELU is used as the activation function for both modules. The dual-view learning objective combines MSE loss for reconstruction and cosine similarity for alignment, and it requires no extra hyperparameters. The hyperparameter settings for the experiment are shown in Table~\ref{tab:Hyperparameter}.


\subsection{Performance Comparison}

As shown in Table \ref{tab:performance}, traditional machine learning methods including SVM and MLP achieve relatively weak overall performance across all metrics. These methods lack the ability to model complex pairwise relationships between subjects, which limits their effectiveness for brain disease diagnosis tasks. Methods built upon brain graphs deliver better results than classic machine learning approaches, yet they still cannot fully exploit cross-subject information for accurate prediction.

Recent population-graph methods generally outperform earlier solutions, as they explicitly model connections among different subjects and make full use of population-level data. Among these competitors, MM-GTUNets obtains competitive results and ranks second on most evaluation metrics. Even so, our PopDRP achieves the best performance in most cases on both datasets. It yields higher accuracy, sensitivity and AUC values than all compared methods, and maintains a favorable specificity level. The consistent improvements demonstrate that the designed graph refinement and multi-scale representation learning modules effectively capture reliable population-level patterns, leading to more robust and accurate diagnosis.

\subsection{Ablation Study}
\begin{table}[!ht]
  \centering
  \caption{Ablation study on ADHD-200 and ABIDE (\%).}
  \label{tab:ablation}
   \resizebox{\linewidth}{!}{
  \begin{tabular}{l c c c c}
    \toprule
    \multirow{2}{*}{Variant} & \multicolumn{2}{c}{ADHD-200} & \multicolumn{2}{c}{ABIDE} \\
    \cmidrule(lr){2-3} \cmidrule(lr){4-5}
    & ACC & AUC & ACC & AUC \\
    \midrule
    Full PopDRP & \textbf{85.79 (1.11)} & \textbf{92.37 (1.22)} & \textbf{84.18 (0.66)} & \textbf{90.55 (0.61)} \\
    \midrule
    w/o NGR & 82.38 (1.02) & 90.18 (1.26) & 80.11 (0.67) & 86.25 (0.59) \\
    w/o CAR & 81.57 (1.40) & 89.42 (1.35) & 79.03 (0.68) & 85.84 (0.67) \\
    w/o SPSS & 72.54 (0.00) & 71.30 (0.68) & 70.97 (0.13) & 71.81 (0.39) \\
    w/o VAE & 69.61 (0.99) & 88.98 (1.25) & 80.73 (0.52) & 87.33 (0.57) \\
    w/o Graph Loss & 80.73 (1.18) & 88.48 (1.34) & 80.80 (0.58) & 87.79 (0.57) \\
    \bottomrule
  \end{tabular}}
\end{table}

We conduct ablation experiments on the ADHD-200 and ABIDE datasets to quantify the contribution of each key component in our PopDRP framework. The detailed results are presented in Table \ref{tab:ablation}.

We first remove the NGR module and keep all other components unchanged. The model achieves lower accuracy and AUC on both datasets, which verifies that neural graph rectification effectively eliminates spurious edges and improves the quality of learned graph structures. When the CAR module is replaced by simple averaging over non-imaging similarities of individual attributes, overall performance drops noticeably. This outcome indicates that contrastive affinity refinement can produce more reliable graph priors compared with naive aggregation strategies.

To examine the effect of the SPSS backbone, we substitute it with a standard graph transformer. The resulting performance experiences a substantial decline, demonstrating that the proposed selective pyramid state-space design is essential for capturing multi-scale topological features and long-range node dependencies. We also test the model by directly feeding raw non-imaging features instead of embeddings learned by VAE. The degraded performance shows that the VAE module learns denoised and compact representations that benefit downstream diagnosis.

Finally, we remove the graph loss term from the overall optimization objective. The slight performance reduction confirms that this regularization term helps constrain graph learning and further enhances model robustness. Compared with all simplified variants, the complete PopDRP system obtains the best results across all metrics, proving the necessity and mutual complementarity of all designed modules.

\begin{figure}[t]
  \centering
  \includegraphics[width=\linewidth]{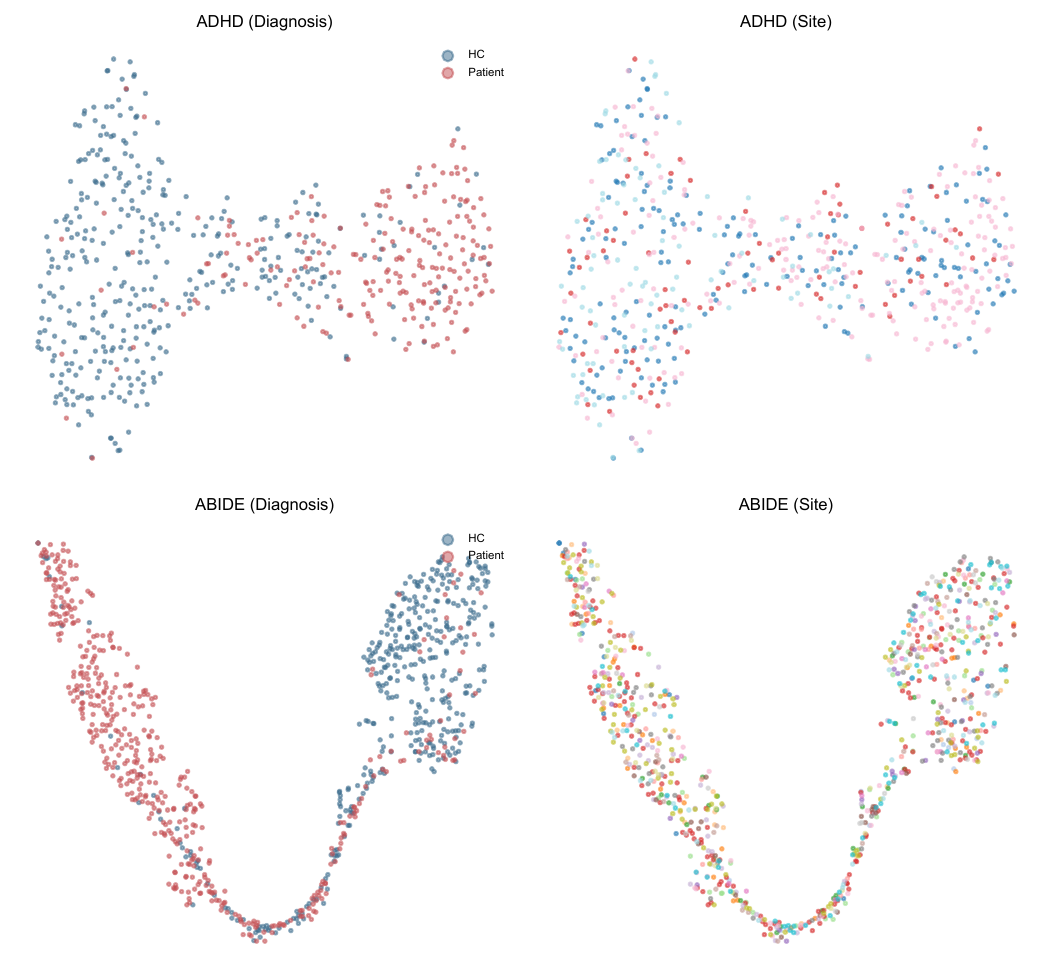}
  \caption{t-SNE visualization of SPSS output embeddings. Diagnosis-colored panels show clear HC/Patient separation; site-colored panels show intermingled site labels, indicating partial mitigation of site effects.}
  \label{fig:tsne}
\end{figure}

\subsection{Hyperparameter Sensitivity}

\begin{figure*}[t]
  \centering
  \includegraphics[width=\textwidth]{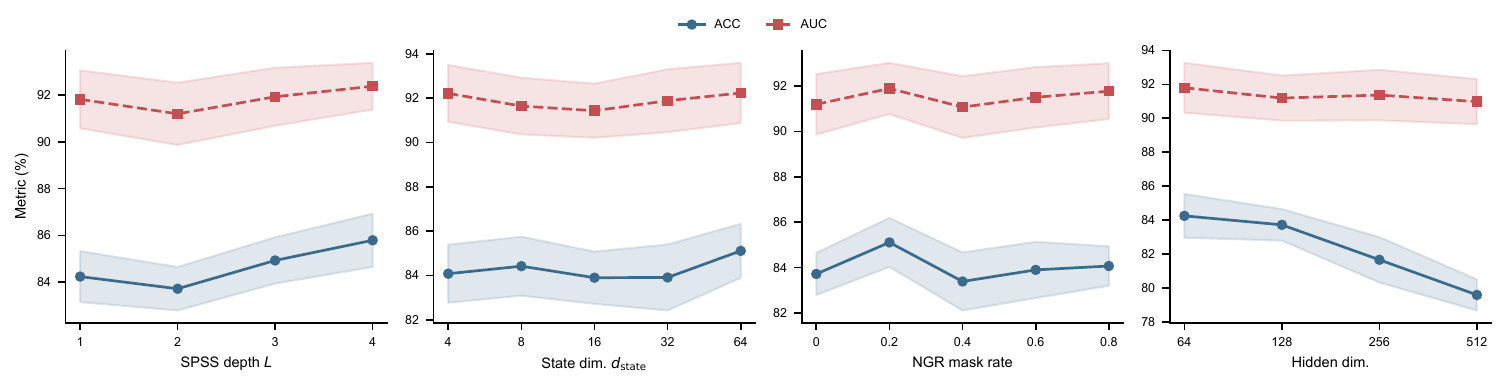}
  \includegraphics[width=\textwidth]{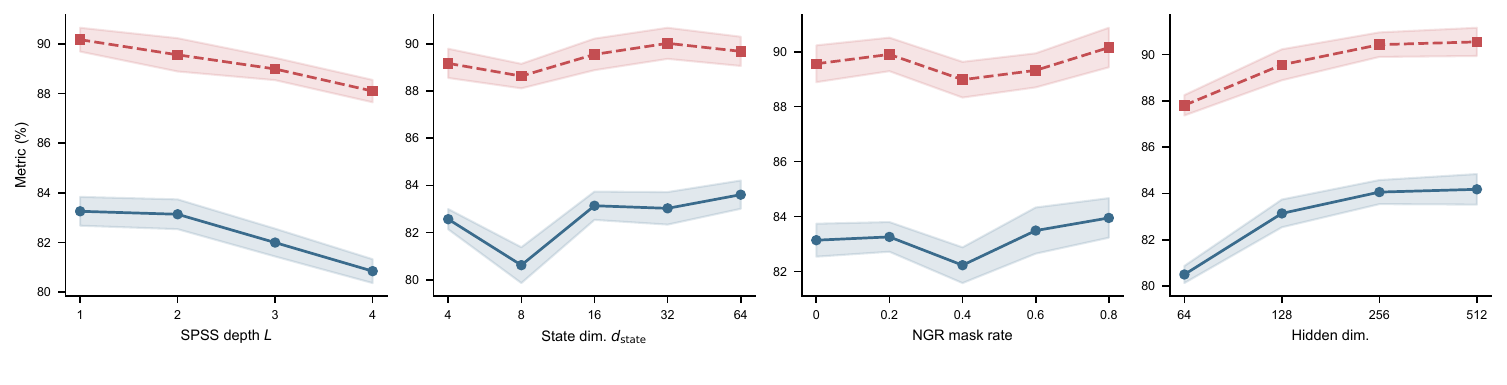}
  \caption{Hyperparameter sensitivity analysis under 10-fold cross-validation  (upper: ADHD-200, lower: ABIDE).  Solid lines and shaded bands denote ACC; dashed lines denote AUC. ADHD and ABIDE exhibit opposite trends for depth and hidden dimension, reflecting their different site heterogeneity profiles.}
  \label{fig:hparam}
\end{figure*}

We conducted a systematic hyperparameter sensitivity analysis on four key parameters: SPSS depth $L$, state dimension $d_{\mathrm{state}}$, NGR mask rate, and hidden dimension. Each parameter was swept independently while holding all others at their default values, using 10-fold cross-validation on both datasets. Figure~\ref{fig:hparam} presents the results for ADHD-200 and ABIDE separately.

The analysis reveals dataset-specific optimal configurations. On ADHD-200, accuracy increases monotonically with SPSS depth, from 84.2\% at $L=1$ to 85.8\% at $L=4$, suggesting that deeper state-space pyramids benefit datasets with lower site heterogeneity. The state dimension $d_{\mathrm{state}}=64$ provides the best accuracy-efficiency trade-off (85.1\%), while smaller hidden dimensions (64) outperform larger ones (84.3\% vs.\ 79.6\% at 512), indicating that ADHD's relatively homogeneous acquisition does not require excessive model capacity. Mask rates of 0.2 provide a modest 0.5 pp gain over the no-mask baseline.

In striking contrast, ABIDE exhibits the opposite trend for SPSS depth: accuracy decreases from 83.3\% at $L=1$ to 80.8\% at $L=4$, indicating that deeper pyramid pooling destroys diagnostically valuable cross-site signal in the highly heterogeneous 17-site dataset. Larger hidden dimensions (512) yield the best ABIDE performance (84.2\%), and higher mask rates (0.6--0.8) provide 0.6--0.8 pp gains, consistent with the hypothesis that stronger regularization benefits multi-site robustness. State dimension $d_{\mathrm{state}}=64$ is the only parameter whose optimal value is consistent across both datasets.


\begin{figure*}[t]
  \centering
  \subfigure[ADHD-200: connections predominantly involve prefrontal-striatal-cingulate circuits.]
  {\includegraphics[width=0.48\textwidth]{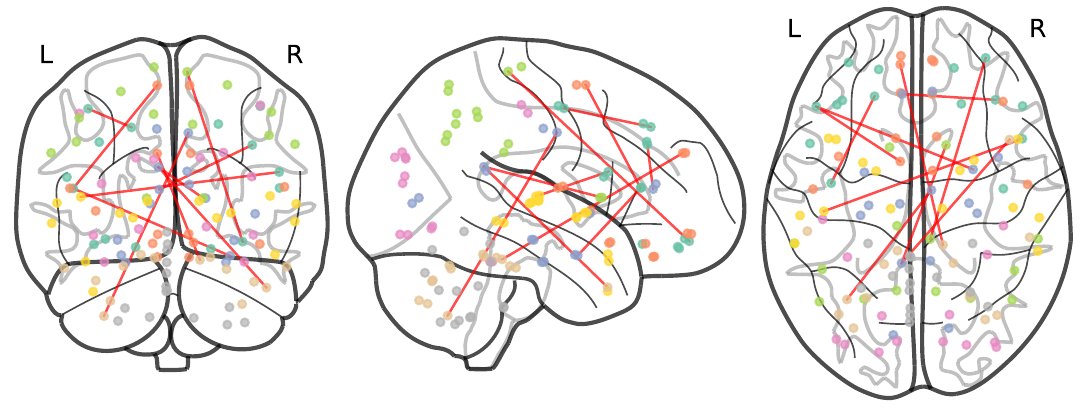}}
  \hfill
  \subfigure[ABIDE: connections are dominated by cerebellar and limbic pathways.]
  {\includegraphics[width=0.48\textwidth]{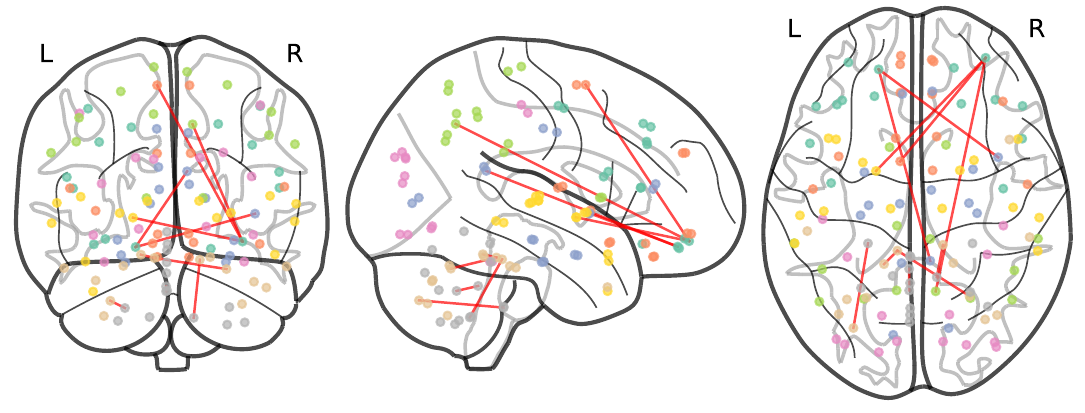}}
  
  \caption{Top-10 PopDRP-selected functional connectivity features.}
  \label{fig:brain}
\end{figure*}

\subsection{Visualization and Interpretability}

We provide four complementary visualizations to assess whether PopDRP learns clinically meaningful population structures and to validate the RFE-selected features against established neuroscience findings.


\textbf{PopDRP-selected brain connectome.} Figure~\ref{fig:brain} shows the top-10 functional connectivity features identified by PopDRP on the glass brain for both datasets. On ADHD-200 (Figure~\ref{fig:brain}a) the selected features concentrate on prefrontal, striatal and cingulate structures---the list contains IFG--amygdala, IFG--SMA and MFG--anterior-cingulate pairs, along with a cingulate--caudate connection, which is the classic fronto-striatal ADHD profile. These fronto-striatal and fronto-limbic pathways are well-established biomarkers of ADHD~\cite{norman2024subcortico}, reflecting the disorder's core deficits in cognitive control, impulse regulation, and emotion processing. On ABIDE (Figure~\ref{fig:brain}b) six of the ten features are cerebellar or vermian (two intra-vermis and four intra-cerebellar pairs) and three further involve orbitofrontal cortex, i.e.\ orbitofrontal--pallidum, orbitofrontal--precuneus and orbitofrontal--SMA. Cerebellar structural and functional abnormalities are among the most consistently replicated neuroimaging findings~\cite{assaf2010abnormal} in ASD, and the fronto-striatal (orbitofrontal-pallidal) circuit is centrally implicated in the repetitive behaviors and social cognition deficits characteristic of the disorder. The alignment between PopDRP-selected features and established neuroscience literature confirms that PopDRP's feature selection captures diagnostically and biologically meaningful signal.

\textbf{t-SNE embedding visualization.} Figure~\ref{fig:tsne} presents t-SNE projections of the SPSS output embeddings. Each row corresponds to one dataset, with the left panel colored by diagnosis and the center panel colored by scanning site; the right panel provides a site-color legend. On both datasets, the diagnosis-colored panels show clear separation between healthy controls and patients, confirming that the SPSS backbone learns diagnostically discriminative representations. The site-colored panels reveal that subjects from different sites are intermingled within each diagnostic cluster rather than forming site-specific subclusters.

\subsection{Limitations and Future Work}

Several limitations motivate future research directions. First, while PopDRP currently models two diagnostic categories per dataset, extending to multi-class diagnosis and dimensional symptom severity prediction would increase clinical utility. Second, the population graph is constructed from a fixed set of non-imaging attributes; incorporating dynamic or longitudinal data could further personalize diagnosis. Third,  all performance assessments in this paper are conducted via internal cross-validation. Additional validation strategies such as testing on independent external cohorts may offer more robust insights into the model’s cross-site transferability.

\section{Conclusion}\label{sec:conclusion}

This work proposes PopDRP, a progressive population graph learning framework for brain disease diagnosis following a disentangle–rectify–perceive pipeline. CAR extracts clinically relevant non-imaging similarities free of confounds; NGR eliminates false graph edges via minimal-sufficient representation; the SPSS backbone captures multi-scale hierarchical features by integrating selective state-space modeling and U-Net graph pooling. Evaluated via 10-fold cross-validation on ABIDE and ADHD-200, PopDRP attains SOTA diagnostic accuracies, with ablation experiments verifying each module’s efficacy. It further delivers interpretable results including structured embeddings to link model outputs to clinical population patterns, alongside biologically meaningful disease-related brain connectivity biomarkers.

\bibliographystyle{IEEEtran}
\bibliography{cas-refs}

@article{gate,
  title={GATE: Graph CCA for temporal self-supervised learning for label-efficient fMRI analysis},
  author={Peng, Liang and Wang, Nan and Xu, Jie and Zhu, Xiaofeng and Li, Xiaoxiao},
  journal={IEEE Transactions on Medical Imaging},
  volume={42},
  number={2},
  pages={391--402},
  year={2022},
  publisher={IEEE}
}

@article{vae,
  title={Auto-encoding variational bayes},
  author={Kingma, Diederik P and Welling, Max},
  journal={arXiv preprint arXiv:1312.6114},
  year={2013}
}

@article{parisot2017spectral,
  title={Spectral graph convolutions for population-based disease prediction},
  author={Parisot, Sarah and Ktena, Sofia Ira and Ferrante, Enzo and Lee, Matthew and Guerrero, Ricardo and Glocker, Ben and Rueckert, Daniel},
  journal={Medical Image Computing and Computer-Assisted Intervention (MICCAI)},
  pages={177--185},
  year={2017},
  publisher={Springer}
}

@article{parisot2018disease,
  title={Disease prediction using graph convolutional networks: Application to autism spectrum disorder and Alzheimer's disease},
  author={Parisot, Sarah and Ktena, Sofia Ira and Ferrante, Enzo and Lee, Matthew and Guerrero, Ricardo and Glocker, Ben and Rueckert, Daniel},
  journal={Medical Image Analysis},
  volume={48},
  pages={117--130},
  year={2018},
  publisher={Elsevier}
}

@article{di2014autism,
  title={The autism brain imaging data exchange: towards a large-scale evaluation of the intrinsic brain architecture in autism},
  author={Di Martino, Adriana and Yan, Chao-Gan and Li, Qingyang and Denio, Erin and Castellanos, Francisco X and Alaerts, Kaat and Anderson, Jeffrey S and Assaf, Michal and Bookheimer, Susan Y and Dapretto, Mirella and others},
  journal={Molecular Psychiatry},
  volume={19},
  number={6},
  pages={659--667},
  year={2014},
  publisher={Nature Publishing Group}
}

@article{bellec2017adhd200,
  title={The ADHD-200 consortium: A model to advance the translational potential of neuroimaging in clinical neuroscience},
  author={Bellec, Pierre and Chu, Carlton and Chouinard-Decorte, Francois and Benhajali, Yassine and Margulies, Daniel S and Craddock, R Cameron},
  journal={Frontiers in Systems Neuroscience},
  volume={11},
  pages={62},
  year={2017},
  publisher={Frontiers Media SA}
}

@article{gu2022efficiently,
  title={Efficiently modeling long sequences with structured state spaces},
  author={Gu, Albert and Goel, Karan and R{\'e}, Christopher},
  journal={International Conference on Learning Representations (ICLR)},
  year={2022}
}

@article{gu2023mamba,
  title={Mamba: Linear-time sequence modeling with selective state spaces},
  author={Gu, Albert and Dao, Tri},
  journal={arXiv preprint arXiv:2312.00752},
  year={2023}
}

@article{wang2024graphmamba,
  title={Graph-Mamba: Towards long-range graph sequence modeling with selective state spaces},
  author={Wang, Chloe and Tongo, Oleksandr and others},
  journal={arXiv preprint arXiv:2402.00789},
  year={2024}
}

@inproceedings{behrouz2024mambaw,
  title={Graph mamba: Towards learning on graphs with state space models},
  author={Behrouz, Ali and Hashemi, Farnoosh},
  booktitle={Proceedings of the 30th ACM SIGKDD conference on knowledge discovery and data mining},
  pages={119--130},
  year={2024}
}

@article{kazi2019inception,
  title={InceptionGCN: receptive field aware graph convolutional network for disease prediction},
  author={Kazi, Anees and Shekarforoush, Shayan and Krishna, S Arvind and Burwinkel, Hendrik and Vivar, Gerome and Kort{\"u}m, Karsten and Ahmadi, Seyed-Ahmad and Navab, Nassir},
  journal={International Conference on Information Processing in Medical Imaging (IPMI)},
  pages={73--85},
  year={2019},
  publisher={Springer}
}

@article{zhu2021deep,
  title={A survey on deep graph structure learning},
  author={Zhu, Yanqiao and Xu, Weizhi and Zhang, Jinghao and Liu, Qiangyu and Wu, Shu and Wang, Liang},
  journal={IEEE Transactions on Neural Networks and Learning Systems},
  year={2021}
}

@article{zheng2023agcl,
  title={A-GCL: Adversarial graph contrastive learning for fMRI analysis to diagnose neurodevelopmental disorders},
  author={Zhang, Shengjie and Chen, Xiang and Shen, Xin and Ren, Bohan and Yu, Ziqi and Yang, Haibo and Jiang, Xi and Shen, Dinggang and Zhou, Yuan and Zhang, Xiao-Yong},
  journal={Medical Image Analysis},
  volume={90},
  pages={102932},
  year={2023},
  publisher={Elsevier}
}

@article{chen2022alnegat,
  title={Adversarial learning based node-edge graph attention networks for autism spectrum disorder identification},
  author={Chen, Yuzhong and Yan, Jiadong and Jiang, Mingxin and others},
  journal={IEEE Transactions on Neural Networks and Learning Systems},
  volume={35},
  number={6},
  pages={7275--7286},
  year={2022}
}

@article{guo2022,
  title={Graph-based fusion of imaging, genetic and clinical data for degenerative disease diagnosis},
  author={Guo, Rui and Tian, Xu and Lin, Hanhe and McKenna, Stephen and Li, Hong-Dong and Guo, Fei and Liu, Jin},
  journal={IEEE/ACM Transactions on Computational Biology and Bioinformatics},
  volume={21},
  number={1},
  pages={57--68},
  year={2023},
  publisher={IEEE}
}

@article{cui2022dgcn,
  author       = {Kanhao Zhao and
                  Boris Duka and
                  Hua Xie and
                  Desmond J. Oathes and
                  Vince D. Calhoun and
                  Yu Zhang},
  title        = {A dynamic graph convolutional neural network framework reveals new
                  insights into connectome dysfunctions in {ADHD}},
  journal      = {NeuroImage},
  volume       = {246},
  pages        = {118774},
  year         = {2022},
  url          = {https://doi.org/10.1016/j.neuroimage.2021.118774},
  doi          = {10.1016/J.NEUROIMAGE.2021.118774},
  bibsource    = {dblp computer science bibliography, https://dblp.org}
}

@article{yao2024mmgl,
  title={Multi-modal graph learning for disease prediction},
  author={Zheng, Shuai and Zhu, Zhenfeng and Liu, Zhizhe and Guo, Zhenyu and Liu, Yang and Yang, Yuchen and Zhao, Yao},
  journal={IEEE Transactions on Medical Imaging},
  volume={41},
  number={9},
  pages={2207--2216},
  year={2022},
  publisher={IEEE}
}

@article{yao2024mmgtunets,
  title={MM-GTUNets: Unified multi-modal graph deep learning for brain disorders prediction},
  author={Cai, Luhui and Zeng, Weiming and Chen, Hongyu and Zhang, Hua and Li, Yueyang and Feng, Yu and Yan, Hongjie and Bian, Lingbin and Siok, Wai Ting and Wang, Nizhuan},
  journal={IEEE Transactions on Medical Imaging},
  year={2025},
  publisher={IEEE}
}

@article{salehi2020gate,
  title={GATE: Graph CCA for temporal self-supervised learning for label-efficient fMRI analysis},
  author={Peng, Liang and Wang, Nan and Xu, Jie and Zhu, Xiaofeng and Li, Xiaoxiao},
  journal={IEEE Transactions on Medical Imaging},
  volume={42},
  number={2},
  pages={391--402},
  year={2022},
  publisher={IEEE}
}

@article{li2022braingnn,
  title={Brain-GNN: Interpretable brain graph neural network for fMRI analysis},
  author={Li, Xiaoxiao and Zhou, Yuan and Dvornek, Nicha C and Zhang, Muhan and Gao, Siyuan and Zhuang, Juntang and Scheinost, Dustin and Staib, Lawrence H and Ventola, Pamela and Duncan, James S},
  journal={Medical Image Analysis},
  volume={74},
  pages={102233},
  year={2021},
  publisher={Elsevier}
}

@article{wang2023evgcn,
  title={Disease prediction with edge-variational graph convolutional networks},
  author={Huang, Yongxiang and Chung, Albert CS},
  journal={Medical Image Analysis},
  volume={77},
  pages={102375},
  year={2022},
  publisher={Elsevier}
}

@inproceedings{song2023rge,
  title={Rge: A repulsive graph rectification for node classification via influence},
  author={Song, Jaeyun and Kim, SungYub and Yang, Eunho},
  booktitle={International Conference on Machine Learning},
  pages={32331--32348},
  year={2023},
  organization={PMLR}
}

@inproceedings{song2022tam,
  title={TAM: topology-aware margin loss for class-imbalanced node classification},
  author={Song, Jaeyun and Park, Joonhyung and Yang, Eunho},
  booktitle={International conference on machine learning},
  pages={20369--20383},
  year={2022},
  organization={PMLR}
}

@article{assaf2010abnormal,
  title={Abnormal functional connectivity of default mode sub-networks in autism spectrum disorder patients},
  author={Assaf, Michal and Jagannathan, Kanchana and Calhoun, Vince D and Miller, Laura and Stevens, Michael C and Sahl, Robert and O'Boyle, Jacqueline G and Schultz, Robert T and Pearlson, Godfrey D},
  journal={Neuroimage},
  volume={53},
  number={1},
  pages={247--256},
  year={2010},
  publisher={Elsevier}
}

@article{norman2024subcortico,
  title={Subcortico-cortical dysconnectivity in ADHD: a voxel-wise mega-analysis across multiple cohorts},
  author={Norman, Luke J and Sudre, Gustavo and Price, Jolie and Shaw, Philip},
  journal={American Journal of Psychiatry},
  volume={181},
  number={6},
  pages={553--562},
  year={2024},
  publisher={American Psychiatric Association Washington, DC}
}

\end{document}